\documentclass[10pt,leqno,twoside]{article}

\usepackage{annal-latex}
\usepackage{amssymb}
\usepackage{graphicx}
\usepackage{algorithm}
\usepackage{algpseudocode}

\newtheorem{theorem}{\indent Theorem}[section]

\newtheorem{lemma}{\indent Lemma}[section]

\newtheorem{definition}{\indent Definition}[section]

\begin{document}
\setcounter{page}{1}
\newcommand\balline{\small Thanh The Van  and Thanh Manh Le}
\newcommand\jobbline{\small Image Retrieval Based on Binary Signature and S-\textit{k}Graph}

\vspace{-4cm} \fofej{43}{14}{105}{122}

\vspace{.4cm}

\title{IMAGE RETRIEVAL BASED ON BINARY SIGNATURE AND S-\textit{k}GRAPH}

\author{{\bf Thanh The Van} (Hue, Vietnam)\\[1ex]
{\bf Thanh Manh Le} (Hue, Vietnam)}

%
\acmclass{H.2.8, H.3.3}
%

%



\commby{J´anos Demetrovics}
\vspace{-2ex}
\recacc{June 1, 2014}{July 1, 2014}

\vspace{-7ex}

\abstract {In this paper, we introduce an optimum approach for querying similar images on large digital-image databases. Our work is based on RBIR (region-based image retrieval) method which uses multiple regions as the key to retrieval images. This method significantly improves the accuracy of queries. However, this also increases the cost of computing. To reduce this expensive computational cost, we implement binary signature encoder which maps an image to its identification in binary. In order to fasten the lookup, binary signatures of images are classified by the help of S-\textit{k}Graph. Finally, our work is evaluated on COREL's images.}

\section{Introduction}
There are three common ways to approach to image retrieval [1], including: text-based image retrieval (TBIR), content-based image retrieval (CBIR) and semantic-based image retrieval (SBIR). The text-based image retrieval is difficult and time-consuming to describe image's content. Thus, it is necessary to build a retrieval system through content of image to find out similarity images. Furthermore, when querying an image through a key word or an index, the features of images can not describe visually. So we need to create a method of extracting image's features to find out images with similarity content. Extracting visual features of image is an important task of image retrieval process based on content. However, if we retrieve and compare directly the content of image, then the problem is complicated, time-consuming and costly storage space. For this reason, when comparing the image's content, we should notice in the query speed and storage space.

A number of works related to the query image's content have been published recently, such as Extracting image objects based on the change of histogram value [1], Similarity image retrieval based on the comparison of characteristic regions and the similarity relationship of feature regions on images [2], Color image retrieval based on the detection of local feature regions by Harris-Laplace [3], Color image retrieval based on bit plane and $L^*a^*b^*$ color space [4], Converting color space and building hash table in order to query the content of color images [5], the similarity of the images based on the combination of the image's colors and texture [9], using the EMD distance in image retrieval [10], the image indexing and retrieval technique VBA (Variable-Bin Allocation) basing on signature bit strings and S-tree [11], etc.

However, if the method of comparing the similarity of the content is ineffective, the results of querying are the images with content not related to the requested query. The approach of the paper is to create the binary signature of an image. The content of the paper aims to query efficiently "\textit{similarity images}" in a large image database system.

The paper approaches the semantic description of image's content through a binary signature and builds a data structure to store binary signatures. This data structure presents the relationship among the binary signatures as well as image's contents. Basing on the description of the semantic relationship of this data structure, the paper finds out the similarity image in content on COREL's image database [6]. The paper contributes two main sections that reduce the amount of query storage and speed up image query on the large image database.

\textbf{The problem:} \textit{Given an image database $\Im $. With each image $J \in \Im $, extract the feature region vector ${R^J} = (R_1^J,R_2^J,...,R_{{n_J}}^J)$ to describe the visual feature of image. Each feature region $R_i^J$ is described as a binary signature $Sig(R_i^J)$. Each query image $I$ is extracted vector of feature region ${R^I} = (R_1^I,R_2^I,...,R_{nI}^I)$ which is described as binary signature $Sig({R^I}) = \bigcup {Sig(R_i^I)} $. Let $\phi (I,J) = \phi ({R^I},{R^J}) = g(Sig({R^I}),Sig({R^J}))$ be a similarity function between image $I$ and $J$. For this reason, with each query image $I$ we need to determine a set of image $Q \subset \Im $ which has the order relation on the base of similarity measure $\phi $.}

To solve the problem, we build a measure which is used to assess the similarity between two images and it is called similarity measure. Basing on this similarity measure, one order set of similarity images which corresponds to query image is selected. At the same time, basing on this order relation, the graph data structure S-\textit{k}Graph is built to describe the similarity relationship in the contents of images. On the base of the data structure, the paper proposes an algorithm which creates S-\textit{k}Graph and a similarity image retrieval algorithm on S-\textit{k}Graph. In order to illustrate the basic theory, the paper gives experiment on a set of COREL images.

The contribution of the paper is an approach to the semantic description of image's content through binary signature as well as building a data structure to store this binary signature. The data structure shows a relationship in the binary signatures which describes the relationship among the contents of images. Basing on the description of semantic relationship of this data structure, the paper finds out similarity images which are conformable to content on COREL image database. [6]

The paper is organized as follows: \textbf{Section 1}, Introduction. \textbf{Section 2}, Presenting construction of theory basis of image's binary signature, the similarity measure between images. \textbf{Section 3}, presenting data structure and image retrieval algorithm based on S-\textit{k}Graph. \textbf{Section 4}, describing the application and assessing the experimental results of the process of finding similarity images. A conclusion and discussion of future works are given in \textbf{Section 5}.

\section{The similarity measure}
According to [7], the binary signature is formed by hashing the data objects, and it has $k$ bits $1$ and $(m-k)$ bits $0$ in the bit chain $[1..m{\rm{]}}$, where $m$ is the length of the binary signature. Data objects and object of the query are encoded on the same algorithm. When the bits in the data object signature are completely covered with the bits in the query signature, then this data object is a candidate of the query. There are three cases: (1) the data object matches the query: each bit in the ${s_q}$ is covered with the bit in the signature ${s_i}$ of the data object (i.e., ${s_q} \wedge {s_i} = {s_q}$); (2) the object does not match the query (i.e., ${s_q} \wedge {s_i} \ne {s_q}$); (3) the signatures are compared and then give a \textit{false drop} result.

In order to evaluate the similarity between two images, firstly the paper builds the binary signature to describe the visual features of each image. On the base of this binary signature, the paper builds similarity measure between two images. The binary signature $Sig(I)$ of the image $I$ is defined as follows:
\begin{definition}
Let $F = ({F_1},...,{F_{{n_F}}})$ be a vector to describe the feature values of region $R_i^I$ of image. Let $F(R_i^I) = ({F_1}(R_i^I),...,{F_{{n_F}}}(R_i^I))$ be a vector value of region feature attribute which is standardized on ${\rm{[}}0,1]$ (i.e: ${F_j}(R_i^I) \in {\rm{[}}0,1],\sum\limits_j {{F_j}(R_i^I) = 1}$, $j = 1,...,{n_F}$). We set $B_I^j = b_1^jb_2^j...b_m^j$ with $b_k^j = 1$ if $k = \left[ {{F_j}(R_i^I)\times m} \right]$, otherwise $b_k^j = 0$, $k = 1,...,m$. At that time, the binary signature of feature region $R_i^I$ is defined as $Sig(R_i^I) = B_I^1B_I^2...B_I^{{n_F}}$. The binary signature of image $I$ is $Sig(I) = Sig({R^I}) = \bigcup\limits_i {Sig(R_i^I)}$.
\end{definition}
In order to increase the accuracy of image query corresponding to the matching feature regions, we need to match the positions of feature regions between the images. For this reason, we need to determine the center positions of feature regions to match the similarity between the images. The center positions of feature regions is defined as follows: 

\begin{definition}
Let ${R^I} = (R_1^I,R_2^I,...,R_{nI}^I)$ be a vector of feature region of image $I(x,y)$. Then, each feature region $R_i^I \in {R^I}$ with center as $C(R_i^I) = ({x_0},{y_0}) = ({{({x_s} - {x_e})} \mathord{\left/
 {\vphantom {{({x_s} - {x_e})} 2}} \right.
 \kern-\nulldelimiterspace} 2},{{({y_s} - {y_e})} \mathord{\left/
 {\vphantom {{({y_s} - {y_e})} 2}} \right.
 \kern-\nulldelimiterspace} 2})$, where ${d_E}(({x_s},{y_s}),({x_e},{y_e})) = \max \{{d_E}(({x_{\alpha i}},\\{y_{\alpha i}}),({x_{\alpha j}},{y_{\alpha j}}))|({x_{\alpha *}},{y_{\alpha *}}) \in Boundary(R_i^I)\} $, with ${d_E}$ as an Euclidean distance and $Boundary(R_i^I)$ as a boundary of feature region $R_i^I$.
\end{definition}
On the base of binary signature and center of feature regions, we set $R_i^I$ and $R_j^J$ in turn as the feature regions on the image $I$ and $J$, respectively. At that moment, the distance between two feature regions is defined as follows: 

\begin{definition}
Let ${R^I} = (R_1^I,R_2^I,...,R_{nI}^I)$ and ${R^J} = (R_1^J,R_2^J,...,R_{nJ}^J)$ be two vectors of feature regions of two images $I(x,y)$ and $J(x,y)$. The distance between feature regions $R_i^I \in {R^I}$ and $R_j^J \in {R^J}$ is $\delta (R_i^I,R_j^J) = ||Sig(R_i^I) - Sig(R_j^J)|{|_1} + {d_E}(C(R_i^I),C(R_j^J))$.
\end{definition}
In order to evaluate the correlation between the measures of images, the following theorem shows that the distance $\delta (R_i^I,R_j^J)$ as a metric.
\begin{theorem}
If ${R^I} = (R_1^I,R_2^I,...,R_{nI}^I)$ and ${R^J} = (R_1^J,R_2^J,...,R_{nJ}^J)$ are two vectors of feature regions of two images $I(x,y)$ and $J(x,y)$ then the distance $\delta (R_i^I,R_j^J)$ is a metric.
\end{theorem}
\proof
(1) Suppose that $R_i^I$ and $R_j^J$ are two feature regions of ${R^I}$ and ${R^J}$. Then, $|Sig(R_i^I) - Sig(R_j^J)|{|_1} \ge 0$ and ${d_E}(C(R_i^I),C(R_j^J)) \ge 0$. Thus, $\delta (R_i^I,R_j^J) = ||Sig(R_i^I) - Sig(R_j^J)|{|_1} + {d_E}(C(R_i^I),C(R_j^J)) \ge 0$. Assume that $\delta (R_i^I,R_j^J) = ||Sig(R_i^I) - Sig(R_j^J)|{|_1} + {d_E}(C(R_i^I),C(R_j^J)) = 0$, then $|Sig(R_i^I) - Sig(R_j^J)|{|_1} = 0$ and ${d_E}(C(R_i^I),C(R_j^J)) = 0$. Furthermore, $||.|{|_1}$ and ${d_E}(.,.)$ are the metrics. So, $Sig(R_i^I) = Sig(R_j^J)$ and $C(R_i^I) = C(R_j^J)$.
Infer, $\delta (R_i^I,R_j^J) \ge 0$ and $\delta (R_i^I,R_j^J) = 0 \Leftrightarrow R_i^I = R_j^J$.
\\(2) Let $k$ be a real number, then: 
\\$\delta (kR_i^I,kR_j^J) = ||Sig(kR_i^I) - Sig(kR_j^J)|{|_1} + {d_E}(C(kR_i^I),C(kR_j^J))$ $ = \sum\limits_{t = 1}^{{n_F}} {|kb_t^I - kb_t^J|}  + \sqrt {{{(kx_0^I - kx_0^J)}^2} + {{(ky_0^I - ky_0^J)}^2}} $ $ 
= |k|\sum\limits_{t = 1}^{{n_F}} {|b_t^I - b_t^J|}  + \sqrt {{k^2}{{(x_0^I - x_0^J)}^2} + {k^2}{{(y_0^I - y_0^J)}^2}} $ $ = |k|\sum\limits_{t = 1}^{{n_F}} {|b_t^I - b_t^J|}  + |k|\sqrt {{{(x_0^I - x_0^J)}^2} + {{(y_0^I - y_0^J)}^2}} $ $ 
\\= |k|\left( {\sum\limits_{t = 1}^{{n_F}} {|b_t^I - b_t^J|}  + \sqrt {{{(x_0^I - x_0^J)}^2} + {{(y_0^I - y_0^J)}^2}} } \right)$ $ 
\\= |k| \times \left( {||Sig(R_i^I) - Sig(R_j^J)|{|_1} + {d_E}(C(R_i^I),C(R_j^J))} \right)$ ${\rm{ = |k|}} \times \delta (R_i^I,R_j^J)$
\\(3) Let ${R^K} = (R_1^K,R_2^K,...,R_{{n_K}}^K)$ be a vector of feature regions of image $K$, then:
$\delta (R_i^I,R_j^J) + \delta (R_j^J,R_k^K) = \left( {||Sig(R_i^I) - Sig(R_j^J)|{|_1} + {d_E}(C(R_i^I),C(R_j^J))} \right)$ $ + \left( {||Sig(R_j^J) - Sig(R_k^K)|{|_1} + {d_E}(C(R_j^J),C(R_k^K))} \right)$ $ 
\\= \left( {||Sig(R_i^I) - Sig(R_j^J)|{|_1} + ||Sig(R_j^J) - Sig(R_k^K)|{|_1}} \right)
\\ + \left( {{d_E}(C(R_i^I),C(R_j^J)) + {d_E}(C(R_j^J),C(R_k^K))} \right)$ 
\\ $ \ge ||Sig(R_i^I) - Sig(R_k^K)|{|_1} + {d_E}(C(R_i^I),C(R_k^K)) = \delta (R_i^I,R_k^K)$.
\\From (1), (2), (3) infer $\delta (R_i^I,R_j^J)$ is a metric.
\qed

On the base of the similarity between the images, the paper builds the similarity measure between two images. On the base of binary signature and feature regions of image, the similarity measure between two images is defined as follows: 
\begin{definition}
Let ${R^I} = (R_1^I,R_2^I,...,R_{{n_I}}^I)$ and ${R^J} = (R_1^J,R_2^J,...,R_{{n_J}}^J)$ be two vectors of feature regions of two images $I(x,y)$ and $J(x,y)$. The similarity function between two images $I$ and $J$ is defined as $\phi (I,J) = \phi ({R^I},{R^J}) = ||Sig(I) - Sig(J)|| + {d_E}(C(I),C(J))$, $C(I) = {{(1} \mathord{\left/
 {\vphantom {{(1} {{n_I}}}} \right.
 \kern-\nulldelimiterspace} {{n_I}}})\sum\limits_i {C(R_i^I)} $.
\end{definition}
\begin{lemma}
The similarity function $\phi (I,J)$ between two images $I$ and $J$ is a metric.
\end{lemma}
\proof
similar to \textit{Theorem 2.1}
\qed

The process of similarity image retrieval is to find a set of images that has the similar content to query image. On the base of the similarity measure at \textit{Definition 2.4}, with each query image $I$, a set of similarity image $Q$ is defined as follows:
\begin{definition}[\textit{Similarity Image Retrieval}]
Let ${\Re _I} = \{ J_i^I|(J_i^I \in \Im ) \wedge (\phi (I,J_i^I) \le \phi (I,J_j^I) \Leftrightarrow J_i^I \succ J_j^I) \wedge (i \ne j) \wedge (i,j = 1,...,n)\} $ be an order set including the images based on the measure $\phi $. A set of similarity images $Q \subset \Im $ includes $k$ similarity images is mean $Q = \{ {J_i} \in \Im |\phi (I,{J_i}) = \phi ({R^I},{R^J}) \le \theta ({R^I},{R^J}),{\rm{ }}\forall J \in \Im ,i = 1,...,k\} $, with $k = |Q|$ and $\theta ({R^I},{R^J})$ is the threshold of $\phi ({R^I},{R^J})$.
\end{definition}

After querying similarity images based on the similarity measure $\phi $, we need to rank the query result according to the similarity measure with the query image. Therefore, a set of result including similarity images $Q$ must be ranked on the similarity measure $\phi $. Following theorem shows a set of result images $Q$ is an order set.
\begin{theorem}
If $I$ is the query image, then the set of similarity images $Q \subset \Im $ is an order set on the relation $ \succ $.
\end{theorem}
\proof
(1) \textit{Symmetry}: If $I$ is the query image and $J \in Q$ is an any image, then $\phi (I,J) = \phi (I,J)$, i.e satisfy condition $\phi (I,J) \le \phi (I,J)$. Hence, $J \succ J$, i.e $Q$ has the symmetry on $ \succ $.
\\(2) \textit{Antisymmetry}: Let ${J_i},{J_j} \in Q$ and $i \ne j$. Suppose that ${J_i} \succ {J_j}$, i.e $\phi (I,{J_i}) \le \phi (I,{J_j})$. Addition ${J_i} \ne {J_j}$ so $\phi (I,{J_i}) < \phi (I,{J_j})$.  Moreover, according to \textit{Lema 2.1}, $\phi $ is a metric. Correspondingly, we have \textit{not} $\phi (I,{J_j}) \le \phi (I,{J_i})$. So, if ${J_i} \succ {J_j}$, then \textit{not} ${J_j} \succ {J_i}$, i.e $Q$ has an antisymmetry on $ \succ $.
\\(3) \textit{Transitivity}: Let ${J_1},{J_2},{J_3}, \in Q$ be three images corresponding to image query $I$, suppose that ${J_1} \succ {J_2}$ and ${J_2} \succ {J_3}$. i.e $\phi (I,{J_1}) \le \phi (I,{J_2})$ and $\phi (I,{J_2}) \le \phi (I,{J_3})$. Otherwise, pursuant to \textit{Lema 2.1}, $\phi $ is a metric, so $\phi (I,{J_1}) \le \phi (I,{J_3})$.
\\Infer: If  ${J_1} \succ {J_2}$ and ${J_2} \succ {J_3}$ then ${J_1} \succ {J_3}$, i.e $Q$ has transitivity on $ \succ $.
\\From (1), (2), (3) we infer the set of similarity images $Q \subset \Im $ is an order set on the relation $ \succ $.
\qed
\section{The data structure and image retrieval algorithm}
\subsection{The S-\textit{k}Graph}
After creating binary signature and similarity measure between the images, the problem is how to query quickly and reduce the query storage. So, we have to build a data structure to store the binary signatures. We also describe the relationship between the images simultaneously. The paper builds the graph structure to describe the similarity relationship based on the binary signature (\textit{Definition 2.1}) and the similarity measure (\textit{Definition 2.4}). This graph structure is called \textit{signature graph} (SG) with each vertex in the graph including the pair of identification $oi{d_I}$  and signature $si{g_I}$ corresponding to image $I$. The weight between two vertexes is the similarity measure $\phi$. The data structure SG is defined as follows:
\begin{definition}[\textit{Signature Graph}]
The signature graph $SG = \left( {V,{\rm{ }}E} \right)$ is the graph which describes the relationship between the images, where is the set of vertexes $V = \{ \langle oi{d_I},Sig({R^I})\rangle |I \in \Im \} $ and the set of edges $E = \{ \langle I,J\rangle |\phi (I,J) = \phi ({R^I},{R^J}) \le \theta ({R^I},{R^J}),{\rm{ }}\forall I,J \in \Im \} $, where $\theta ({R^I},{R^J})$ is a threshold value and $\Im $ is an image database. The weight of each edge $\langle I,J\rangle $ is a measurement function of the similarity $\phi (I,J) = \phi ({R^I},{R^J})$, 
\end{definition}

Each vertex $v \in V$ in $SG$ determines $k$ elements which has the nearest similar measurement. However, with the number of images in a large database, it is difficult to determine the set of similarity image corresponding to the query image. Therefore, we build the notion of S-\textit{k}Graph so that each vertex includes the nearest image and called \textit{k-neighboring} image.

With each \textit{k-neighboring} image, the paper builds a cluster including similarity images. This cluster represents an item called center cluster. Then, each cluster includes similarity images is defined as follows:
\begin{definition}
A cluster ${V_i}$ has center ${I_i}$, with ${k_i}\theta $ as a radius, is defined as follows: ${V_i} = {V_i}({I_i}) = \{ J|\phi ({I_i},J) \le {k_i}\theta ,J \in \Im ,{\rm{ }}i = 1,...,n\} $, ${k_i} \in {N^*}$.
\end{definition}

On the base of clusters, the paper defines the data structure S-\textit{k}Graph including vertexes as clusters and the weight between two vertexes as the similarity measure $\phi$. The data structure S-\textit{k}Graph is defined as follows:
\begin{definition}[\textit{S-kGraph}]
Let $\Omega  = \{ {V_i}|i = 1,...,n\} $ be a set of clusters so that ${V_i} \cap {V_j} = \emptyset ,i \ne j$. The S-\textit{k}Graph = $\left( {{V_{SG}},{\rm{ }}{E_{SG}}} \right)$ is the graph with the weight, including a vertex set ${V_{SG}}$ and an edge set ${E_{SG}}$ which are defined as follows: ${V_{SG}} = \Omega  = {\rm{\{ }}{V_i}|\exists !{I_{{i_0}}} \in {V_i},\forall I \in {V_i},\phi ({I_{{i_0}}},I) \le {k_{{i_0}}}\theta ,i = 1,...,n{\rm{\} }}$, ${E_{SG}} = \{ \langle {V_i},{V_j}\rangle |i \ne j,{V_i} \in {V_{SG}},{V_j} \in {V_{SG}},d({V_i},{V_j}) = \phi ({I_{{i_0}}},{J_{{j_0}}})\} $, where $d({V_i},{V_j})$ is the weight between two clusters and $\forall I \in {V_i},\phi ({I_{{i_0}}},I) \le {k_{{i_0}}}\theta $.
\end{definition}

With each image we need to classify in clusters through the data structure S-\textit{k}Graph. So, we need to have the rules of distribution in clusters of the S-\textit{k}Graph. These rules are defined as follows:
\begin{definition}[\textit{The Rules of Distribution of Image}]
Let $\Omega  = \{ {V_i}|i = 1,...,n\} $ be a set of clusters so that ${V_i} \cap {V_j} = \emptyset ,i \ne j$, ${I_0}$ be an image which needs to distribute in a set of clusters $\Omega $, ${I_m}$ be a center of cluster ${V_m}$ so that $(\phi ({I_0},{I_m}) - {k_m}\theta ) = \min \{ (\phi ({I_0},{I_i}) - {k_i}\theta ),i = 1,...,n{\rm{\} }}$, where ${I_i}$ is a center of cluster ${V_i}$. There are three cases as follows: 
\\(1) If $\phi ({I_0},{I_m}) \le {k_m}\theta $ then the image ${I_0}$ is distributed in cluster ${V_m}$.
\\(2) If $\phi ({I_0},{I_m}) > {k_m}\theta $ then setting ${k_0} = \left[ {{{(\phi ({I_0},{I_m}) - {k_m}\theta )} \mathord{\left/
 {\vphantom {{(\phi ({I_0},{I_m}) - {k_m}\theta )} \theta }} \right.
 \kern-\nulldelimiterspace} \theta }} \right]$, at that time: 
\\(2.1) If ${k_0} > 0$ then creating cluster ${V_0}$ with center ${I_0}$ and radius ${k_0}\theta $, at that time $\Omega  = \Omega  \cup {\rm{\{ }}{V_0}{\rm{\} }}$.
\\(2.2) Otherwise (i.e ${k_0} = 0$), the image ${I_0}$ is distributed in cluster ${V_m}$ and $\phi ({I_0},{I_m}) = {k_m}\theta $.
\end{definition}

Each image needs to exist a cluster in the S-\textit{k}Graph so that images are classified. Moreover, to avoid the invalid data in clusters, the images are distributed in unique cluster. The \textit{theorem 3.1} and \textit{theorem 3.2} show the unique distribution.

\begin{theorem}
Given the S-\textit{k}Graph = $\left( {{V_{SG}},{\rm{ }}{E_{SG}}} \right)$. Let $\langle {V_i},{V_j}\rangle  \in {E_{SG}}$ and ${I_{{i_0}}},{J_{{j_0}}}$ in turn be a center of ${V_i},{V_j}$. At that time, $d({V_i},{V_j}) = \phi ({I_{{i_0}}},{J_{{j_0}}}) > ({k_{{i_0}}} + {k_{{j_0}}})\theta $, with $\forall I \in {V_i},\phi ({I_{{i_0}}},I) \le {k_{{i_0}}}\theta $ and $\forall J \in {V_j},\phi ({J_{{j_0}}},J) \le {k_{{j_0}}}\theta $.
\end{theorem}
\proof
So $\forall I \in {V_i},\phi ({I_{{i_0}}},I) \le {k_{{i_0}}}\theta $ and $\forall J \in {V_j},\phi ({J_{{j_0}}},J) \le {k_{{j_0}}}\theta $. That $\forall I' \in Boundary({V_i}),\forall J' \in Boundary({V_j})$ then $\phi ({I_{{i_0}}},I') = {k_{{i_0}}}\theta $ and $\phi ({J_{{j_0}}},J') = {k_{{j_0}}}\theta $. Moreover, because ${V_{SG}} = \Omega $ is a set of unconnected cluster, so  ${V_i} \cap {V_j} = \emptyset $ that $\phi (I',J') > 0$.
\\Infer: $\forall I' \in Boundary({V_i}),\forall J' \in Boundary({V_j})$ then $\phi ({I_{{i_0}}},I') + \phi (I',J') + \phi ({J_{{j_0}}},J') > ({k_{{i_0}}} + {k_{{j_0}}})\theta $. Otherwise, because $\phi $ is a metric, so $\phi ({I_{{i_0}}},I') + \phi (I',J') + \phi ({J_{{j_0}}},J') \ge \phi ({I_{{i_0}}},{J_{{j_0}}})$. And $\exists {I'_0} \in Boundary({V_i}),\exists {J'_0} \in Boundary({V_j})$ so as $\phi ({I_{{i_0}}},{I'_0}) + \phi ({I'_0},{J'_0}) + \phi ({J_{{j_0}}},{J'_0}) = \phi ({I_{{i_0}}},{J_{{j_0}}})$.
\\ Therefore, $\phi ({I_{{i_0}}},{J_{{j_0}}}) > ({k_{{i_0}}} + {k_{{j_0}}})\theta $.
\qed

\begin{theorem}
If each image $I$ is distributed in a set of clusters $\Omega  = \{ {V_i}|i = 1,...,n\} $, then it belongs to an unique cluster.
\end{theorem}
\proof
Let $I$ be an any image, suppose that $\exists {V_i},{V_j}$ as two clusters, so ${V_i} \ne {V_j}$ and $(I \in {V_i}) \wedge (I \in {V_j})$. Setting ${I_i},{I_j}$ in turn as two centers cluster ${V_i},{V_j}$ we have $\phi ({I_i},I) \le {k_i}\theta $ and $\phi ({I_j},I) \le {k_j}\theta $. Thus, $\phi ({I_i},I) + \phi ({I_j},I) \le ({k_i} + {k_j})\theta $. Furthermore, because $\phi $ is a metric, we have $\phi ({I_i},I) + \phi ({I_j},I) \ge \phi ({I_i},{I_j})$. Otherwise, ${I_i},{I_j}$ in turn as two centers cluster ${V_i},{V_j}$ so that $\phi ({I_i},{I_j}) > ({k_i} + {k_j})\theta $. Hence, $\phi ({I_i},I) + \phi ({I_j},I) \ge \phi ({I_i},{I_j}) > ({k_i} + {k_j})\theta $ and $\phi ({I_i},I) + \phi ({I_j},I) \le ({k_i} + {k_j})\theta $. 
\\For this reason, the supposition is illogical. I.e each image $I$ is only distributed in an unique cluster.
\qed

In order to avoid invaliding data, the rules of distribution (\textit{Definition 3.4}) needs to ensure that the image is classified in an unique cluster. \textit{Theorem 3.3}, \textit{theorem 3.4} and \textit{theorem 3.5} show this problem.
\begin{theorem}
If the value $\phi (I,{I_m}) - {k_m}\theta  \le 0$ then it only occurs at one unique ${I_m}$.
\end{theorem}
\proof
Suppose that $\exists {I_0}$ is a center of cluster ${C_0} \in \Omega $ so that $\phi (I,{I_0}) - {k_0}\theta  \le 0$ $ \Leftrightarrow $ $\phi (I,{I_0}) \le {k_0}\theta $, i.e $I$ belongs to cluster ${C_0}$. Otherwise, according to the supposition, $\phi (I,{I_m}) - {k_m}\theta  \le 0$, i.e  $I$ belongs to cluster ${C_m} \ne {C_0}$. It means that $I$ belongs to two different clusters and pursues to \textit{Theorem 3.2}, each image $I$ only belongs to an unique cluster. Thus, the supposition is illogical. Inferring, if the value is $\phi (I,{I_m}) - {k_m}\theta  \le 0$, it only occurs at one unique ${I_m}$.
\qed
\begin{theorem}
If $\Omega  = \{ {V_i}|i = 1,...,n\} $ be a set of clusters and $I$ is an image then it exists cluster ${V_{{i_0}}} \in \Omega $ so that $I \in {V_{{i_0}}}$.
\end{theorem}
\proof
According to \textit{Definition 3.4}, any image $I$ also exists a cluster ${V_{{i_0}}} \in \Omega $ so that $I \in {V_{{i_0}}}$.
\qed

\begin{theorem}
Each image $I$ is distributed in an unique cluster ${C_{{i_0}}}\in \Omega$.
\end{theorem}
\proof
According to \textit{Definition 3.4}, any image $I$ also exists a cluster ${V_{{i_0}}} \in \Omega $ so that $I \in {V_{{i_0}}}$. According to \textit{Theorem 3.2}, any image $I$ is only distributed in an unique cluster. Inferring, any image $I$ is distributed in an unique cluster ${C_{{i_0}}} \in \Omega $.
\qed
\subsection{Extracting the feature regions}
In order to execute the similarity image retrieval process according to the proposed theory, we firstly extract the feature regions of the image. The paper presents the method to extract the feature regions based on the interest points on image. This interest points are extracted with the intensity and Harris-Laplace detector. 

In order to extract the visual features of image, the first step is standardized the image size. Let Y, Cb, Cr be Intensity, Blue color, Red color, respectively. According to [3], [4], the Gaussian transformation by human's visual system is fulfilled as follows:
$L(x,y) = \frac{1}{{10}}{\rm{[}}6.G(x,y,{\delta _D})*Y + 2.G(x,y,{\delta _D})*Cb + 2.G(x,y,{\delta _D})*Cr]$
with $G(x,y,{\delta _D}) = \frac{1}{{\sqrt {2\pi } .{\delta _D}}}.\exp (\frac{{{x^2} + {y^2}}}{{2.\delta _D^2}})$. The intensity ${I_0}(x,y)$ for color image is calculated according to equation: ${I_0}(x,y) = Det(M(x,y)) - \alpha .T{r^2}(M(x,y,))$, where $Det( \bullet ),Tr( \bullet )$ are Determinant and Trace of matrix, respectively. $M(x,y)$ is a second moment matrix $M(x,y) = \delta _D^2.G({\delta _I})*\left[ {\begin{array}{*{20}{c}}
{{L_x}^2}&{{L_x}{L_y}}\\
{{L_x}{L_y}}&{{L_y}^2}
\end{array}} \right]$, where ${\delta _I},{\delta _D}$ are the integration scale and differentiation scale, and ${L_\alpha }$ is the derivative computed the $\alpha $ direction. The interest points of color image are extracted according to formula: ${I_0}(x,y) > {I_0}(x',y')$, with $x',y' \in A$, ${I_0}(x,y) \ge \theta $, where $A$ is the neighboring of point $(x,y)$ and $\theta $ is a threshold value. 

Let ${O_I} = {\rm{\{ }}o_I^1,o_I^2,...,o_I^n{\rm{\} }}$ be a set of feature circles with its center as a interest points and a set of feature radius ${R_I} = {\rm{\{ }}r_I^1,r_I^2,...,r_I^n{\rm{\} }}$. Values of feature radius are extracted with LoG method (Laplace-of-Gaussian) and their value in ${\rm{[}}0,{{\min (M,N)} \mathord{\left/
 {\vphantom {{\min (M,N)} 2}} \right.
 \kern-\nulldelimiterspace} 2}{\rm{]}}$, where $M,N$ are the height and the width of image.
\begin{figure}[htbp]
\centering
\includegraphics[width=340px,height=60px]{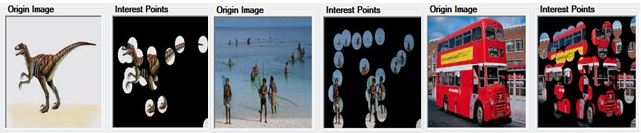}
\caption{A sample result of extracting feature region} \label{fig:graph}
\end{figure}
For each image, the process of extraction interest points is described as follows:

\textit{Step 1.} Convert from RGB color space to YCbCr color space.

\textit{Step 2.} Perform Gaussian transform for the human visual system to calculate the $L(x,y)$.

\textit{Step 3.} Calculate the feature intensity ${I_0}(x,y)$ for color images. Then, collect the set of interest points.

\textit{Step 4.} Implement of the extraction feature regions ${O_I} = {\rm{\{ }}o_I^1,o_I^2,...,o_I^n{\rm{\} }}$ based on the interest points.
\subsection{Binary signature of the image}
After extracting the feature regions of image, we need to create the binary signatures to describe these. On the base of the binary signatures, we perform the similarity image retrieval process for the proposed theory.

With each feature region $o_I^i \in {O_I}$ of the image $I$, the histogram is calculated  on the base of the standard color range $C$. Effective clustering method relies on Euclidean measure in RGB color space classify colors of every pixel on the image. Let $p$ be a pixel of image $I$ which has a color vector in RGB as ${V_p} = \left( {{\rm{ }}{R_p},{\rm{ }}{G_p},{\rm{ }}{B_p}} \right)$, ${V_m} = \left( {{R_m},{\rm{ }}{G_m},{\rm{ }}{B_m}} \right)$ be a color vector of a set of standard color range $C$, so as ${V_m} = {\rm{ }}\min \{ ||{V_p} - {V_i}||,{\rm{ }}{V_i} \in C\} $. At that time, pixel $p$ is standardized in accordance with color vector ${V_m}$. According to experiment, the paper uses the standard color range on MPEG7 to calculate histogram for color images on COREL database. 

Setting $o_I^i \in {O_I}$ ($i = 1,...,N$) a feature circle of the image $I$, the histogram vector of the circle $o_I^i$ is $H(o_I^i) = \{ {H_1}(o_I^i),...,{H_{n}}(o_I^i)\} $. Setting ${h_k}(o_I^i) = \frac{{{H_k}(o_I^i)}}{{\sum\limits_j {{H_j}(o_I^i)} }}$, a standard histogram vector is $h(o_I^i) = \{ {h_1}(o_I^i),...,{h_{n}}(o_I^i)\} $. Then, the binary signature describes ${h_k}(o_I^i)$ as $B_I^k = b_I^1b_I^2...b_I^{m}$, with $b_I^j = 1$ if $j = \left[ {({h_j}(o_I^i) + 0.05) \times m} \right]$, otherwise $b_I^j = 0$. So, the signature describes the feature region $o_I^i \in {O_I}$ as $Sig(o_i^I) = B_I^1B_I^2...B_I^n$. For this reason, the binary signature of the image $I$ is ${S_I} = \bigcup\nolimits_{i = 1}^N {Sig(o_I^i)} $. 
The process of creating binary signatures for color images is described as follows:

\textit{Step 1. }Calculate the histogram vector $H(o_I^i) = \{ {H_1}(o_I^i),{H_2}(o_I^i),...,{H_n}(o_I^i)\} $ on the base of feature region $o_I^i \in {O_I}$ with the set of standard color $C$.

\textit{Step 2.} For each the feature region $o_I^i \in {O_I}$, standardize histogram vector as $h(o_I^i) = \{ {h_1}(o_I^i),{h_2}(o_I^i),...,{h_n}(o_I^i)\} $.

\textit{Step 3.} Create the binary signature for ${h_k}(o_I^i)$ as $B_I^k = b_I^1b_I^2...b_I^m$, with $b_I^j = 1$ if $j = \left[ {({h_j}(o_I^i) + 0.05) \times m} \right]$, otherwise $b_I^j = 0$. The signature describes the feature region $o_I^i \in {O_I}$ as $Sig(o_i^I) = B_I^1B_I^2...B_I^n$.

\textit{Step 4.} Create the binary signature of image $I$ as ${S_I} = \bigcup\nolimits_{i = 1}^N {Sig(o_I^i)} $.
\subsection{Creating S-\textit{k}Graph}
On the base of the similarity measure $\phi$, the S-\textit{k}Graph is shown in \textit{Definition 3.3} and the rules of distribution of image are shown in \textit{Definition 3.4}, the paper proposes the algorithm to create the data structure S-\textit{k}Graph. With the input image database $\Im $ and the threshold $k\theta$, we need to return the S-\textit{k}Graph. Firstly, we initialize the set of vertex ${V_{SG}} = \emptyset $ and initialize the set of edge ${E_{SG}} = \emptyset $, after that create the first cluster. With each image $I$ we evaluate the distance $\phi $ with the center of cluster and to find out the nearest cluster according to $(\phi (I,I_0^m) - {k_m}\theta ) = \min \{ (\phi (I,I_0^i) - {k_i}\theta ),i = 1,...,n{\rm{\} }}$. If the condition $\phi (I,I_0^m) \le {k_m}\theta $ is satisfied, the image $I$ is distributed in cluster ${V_m}$. Otherwise, we consider the rules of distribution as shown in \textit{Definition 3.4} to classify the image $I$ into appropriate cluster. This algorithm is as follows:
\\\textbf{\textit{Algorithm 1.}} Create the S-\textit{k}Graph \\
\textbf{\textit{Input:}} Image database $\Im $ and threshold $k\theta $\\
\textbf{\textit{Output:}} S-\textit{k}Graph = $({V_{SG}},{E_{SG}})$
\begin{algorithmic}[1]
\State ${V_{SG}} = \emptyset;$ ${E_{SG}} = \emptyset ;$ ${k_I} = 1;$ $n = 1;$
\For {$\forall I \in \Im $}
	\If {${V_{SG}} = \emptyset $}
		\State $I_0^n = I;$ $r = {k_I}\theta ;$
		\State Initialize cluster ${V_n} = \langle I_0^n,r,\phi  = 0\rangle ;$
		\State ${V_{SG}} = {V_{SG}} \cup {V_n};$
	\Else
		\State $(\phi (I,I_0^m) - {k_m}\theta ) = \min \{ (\phi (I,I_0^i) - {k_i}\theta ),i = 1,...,n{\rm{\} }}$
		\If {$\phi (I,I_0^m) \le {k_m}\theta $}
			\State ${V_m} = {V_m} \cup \langle I,{k_m}\theta ,\phi (I,I_0^m)\rangle ;$
		\Else 
			\State ${k_I} = \left[ {{{(\phi (I,I_0^m) - {k_m}\theta )} \mathord{\left/
							{\vphantom {{(\phi (I,I_0^m) - {k_m}\theta )} \theta }} \right.
							\kern-\nulldelimiterspace} \theta }} \right];$
			\If {${k_I} > 0$}
				\State $I_0^{n + 1} = I;$ $r = {k_I}\theta ;$
				\State Initialize cluster ${V_{n + 1}} = \langle I_0^{n + 1},r,\phi  = 0\rangle ;$
				\State ${V_{SG}} = {V_{SG}} \cup {V_{n + 1}};$
				\State ${E_{SG}} = {E_{SG}} \cup {\rm{\{ }}\langle {V_{n + 1}},{V_i}\rangle |\phi (I_0^{n + 1},I_0^i) \le k\theta ,i = 1,...,n{\rm{\} }};$
				\State $n = n + 1;$
			\Else
				\State $\phi (I,I_0^m) = {k_m}\theta ;$
				\State ${V_m} = {V_m} \cup \langle I,{k_m}\theta ,\phi (I,I_0^m)\rangle ;$
			\EndIf
		\EndIf
	\EndIf
	\EndFor
	\State \textbf{Return} S-\textit{k}Graph = $({V_{SG}},{E_{SG}});$
\end{algorithmic}
\subsection{Image retrieval algorithm}
After creating the S-\textit{k}Graph, we need to query the similarity images on it. With each query image ${I_Q}$, we need to query the set of the similarity images $IMG$. This query process finds out the nearest cluster in S-\textit{k}Graph with ${\phi _{\min }} = \phi ({I_Q},I_0^m) = \min \{ \phi ({I_Q},I_0^i),i = 1,...,n\} $. On the other hand, we need to query the similarity images at adjacent vertex with the measure less than threshold $k\theta $. This algorithm is described as follows:
\\\textbf{\textit{Algorithm 2.}} Image Retrieval Algorithm based on S-\textit{k}Graph\\
\textbf{\textit{Input:}} query image ${I_Q}$, S-\textit{k}Graph=$({V_{SG}},{E_{SG}})$, threshold $k\theta $\\
\textbf{\textit{Output:}} set of a similarity image $IMG$
\begin{algorithmic}[1]
	\State $IMG = \emptyset ;$ $V = \emptyset ;$
	\State ${\phi _{\min }} = \phi ({I_Q},I_0^m) = \min \{ \phi ({I_Q},I_0^i),i = 1,...,n\} ;$
	\For {${V_i} \in {V_{SG}}$}
		\If {$\phi (I_0^m,I_0^i) \le k\theta $}
			\State $V = V \cup {V_i};$
		\EndIf
	\EndFor
	\For {${V_j} \in V$}
		\State $IMG = IMG \cup \{ I_k^j,I_k^j \in {V_j},k = 1,...,|{V_j}|\} ;$
	\EndFor
	\State\textbf{Return} $IMG;$
\end{algorithmic}
\section{Experiments}
\subsection{Model of image retrieval system}
\begin{figure}[ht]
	\centering
		\includegraphics[height=4cm,width=6cm]{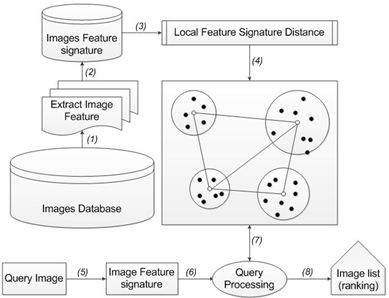}
	\caption{The model of RBIR using S-\textit{k}Graph}
\end{figure}
\begin{figure}[ht]
	\centering
		\includegraphics[height=4.5cm,width=9cm]{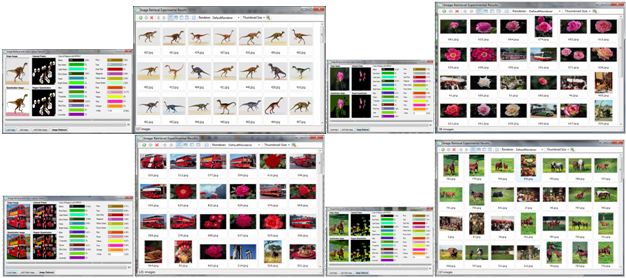}
	\caption{A sample result of image retrieval based on S-\textit{k}Graph}
\end{figure}
\textbf{\textit{Phase 1:}} \textit{Perform pre-processing}

\textit{Step 1.} Extract feature regions of the images in database into the form of feature vectors.

\textit{Step 2.} Convert the feature vectors of the image into the form of binary signatures.

\textit{Step 3.} Calculate the similarity measure among the binary signatures of the images and insert into S-\textit{k}Graph.

\textbf{\textit{Phase 2:}} \textit{Implement Query}

\textit{Step 1.} For each query image, we extract the feature vector and convert into binary signature.

\textit{Step 2.} Perform the process of binary signature retrieval on S-\textit{k}Graph to find out the similarity images.

\textit{Step 3.} After creating the similarity images, we carry out an arrangement from high to low and give a list of the images on the base of the similarity binary signatures.
\subsection{The experimental results}
The experimental processing on COREL sample data [6] including 10,800 images which are divided into 80 different subjects. With each query image, we retrieve images on COREL data as so as find out the most similar ones to the query image. Then, we compare to the list of subjects of images to evaluate the accurate method.

Binary signatures are introduced into two forms of query structure including SSF (sequential signature file) and S-\textit{k}Graph. Fig.6 and Fig.7 describe empirical figures about the similarity image retrieval process on COREL images.
\begin{figure}[!ht]
	\centering
		\includegraphics[height=4.7 cm,width=8.7cm]{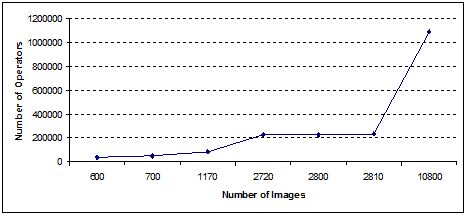}
		\caption{Number of comparisons to create S-\textit{k}Graph}
\end{figure}
\begin{figure}[!ht]
	\centering
		\includegraphics[height=4.7 cm,width=8.7cm]{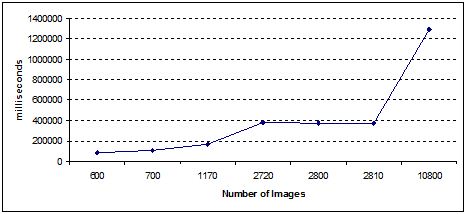}
	\caption{The time to create S-\textit{k}Graph}
\end{figure}
\begin{figure}[!ht]
	\centering
		\includegraphics[height=4.7 cm,width=8.7cm]{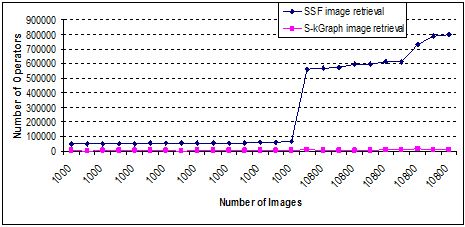}
	\caption{Number of comparisons to query image}
\end{figure}
\begin{figure}[!ht]
	\centering
		\includegraphics[height=4.7 cm,width=8.7cm]{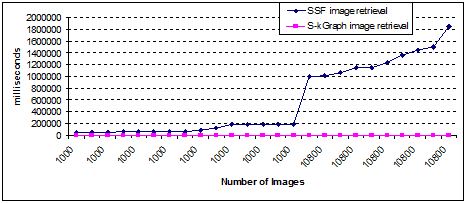}
	\caption{The time to query image}
\end{figure}
\begin{figure}[!ht]
	\centering
		\includegraphics[height=4.7cm,width=8.7cm]{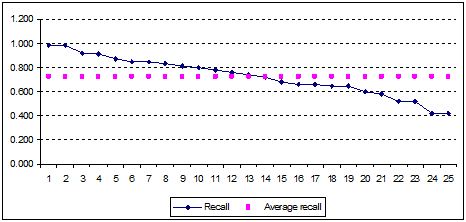}
	\caption{Recall}
\end{figure}
\begin{figure}[!ht]
	\centering
		\includegraphics[height=4.7cm,width=8.7cm]{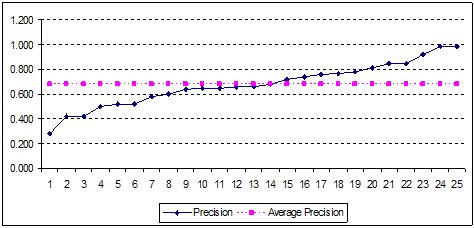}
	\caption{Precision}
\end{figure}
\section{Conclusion}
The paper gives a similar evaluation method between two images on the base of binary signature and creates S-\textit{k}Graph to describe the relationship between images. As a result, the paper creates the image retrieval system model on the base of feature regions which is to simulate the experiment on COREL's image data classification. According to experimental results, the method of evaluation which is based on S-\textit{k}Graph speeds up in query similarity images more than query in SSF (sequential signature file). However, the use of the features of color gives an inaccurate result in the sense of image content. Therefore, the next development is to extract objects on the image. Consequently, the paper gives binary signatures to describe objects as well as the contents of images. On the base of these binary signatures, we assess the similarity measure and return the set of similarity images with query image.

\vspace{2cm}

\noindent\textbf{Thanh The Van}\\
Faculty of Information Technology\\
Hue University of Sciences, Hue University\\
77 Nguyen Hue street\\
Hue city\\
Vietnam\\
{\tt vanthethanh@gmail.com}\\

\noindent\textbf{Thanh Manh Le}\\
Hue University\\
03 Le Loi street\\
Hue city\\
Vietnam\\
{\tt lmthanh@hueuni.edu.vn}

\end{document}